\documentclass[11pt, letterpaper, logo, onecolumn, copyright, numbering]{acc}

\usepackage{url}

\usepackage{natbib}

\usepackage{amsthm}

\usepackage{times}

\definecolor{darkblue}{RGB}{32, 64, 129}
\definecolor{darkgreen}{RGB}{0, 110, 85}
\definecolor{darkred}{RGB}{153, 0, 0}
\definecolor{graytext}{gray}{0.45}

\definecolor{evaluatorcolor}{RGB}{255,230,230}    %
\definecolor{evaluatorframe}{RGB}{180,50,50}      %
\definecolor{taskgencolor}{RGB}{230,255,230}      %
\definecolor{taskgenframe}{RGB}{50,180,50}        %
\definecolor{executioncolor}{RGB}{230,230,255}    %
\definecolor{executionframe}{RGB}{50,50,180}      %
\definecolor{lightgray}{RGB}{240,240,240}
\definecolor{darkgray}{RGB}{80,80,80}

\usepackage{pifont}
\usepackage{longtable}

\usepackage{makecell}

\usepackage{amsmath}
\usepackage{pifont}

\usepackage{booktabs}
\usepackage{multirow}
\usepackage{graphicx}
\usepackage{float}
\usepackage{caption}
\usepackage{subcaption}
\usepackage{xspace}

\usepackage[utf8]{inputenc}
\usepackage[T1]{fontenc}
\usepackage{xcolor}
\usepackage[most]{tcolorbox}
\usepackage{enumitem}
\usepackage{listings}

\definecolor{darkgray}{rgb}{0.3, 0.3, 0.3}
\definecolor{lightgray}{rgb}{0.95, 0.95, 0.95}
\definecolor{codegray}{rgb}{0.98, 0.98, 0.98}

\newtcolorbox{promptbox}[1]{
    colback=lightgray,
    colframe=darkgray,
    colbacktitle=darkgray,
    coltitle=white,
    boxrule=2pt,
    arc=0mm,
    left=10pt,
    right=10pt,
    top=10pt,
    bottom=10pt,
    fonttitle=\bfseries\large,
    title={#1},
    attach boxed title to top left={yshift=-2mm}
}

\usepackage{wrapfig}

\usepackage{algorithm}
\usepackage[noend]{algpseudocode}

\definecolor{deepred}{rgb}{0.631,0.102,0.102}
\definecolor{skyblue}{HTML}{126da2}

\definecolor{accpurple}{HTML}{A100FF}
\hypersetup{
     colorlinks = true,
     breaklinks = true,
     linkcolor = accpurple,
     urlcolor = accpurple,
     citecolor = accpurple
     }

\definecolor{orange}{rgb}{1,0.5,0}

\algnewcommand{\LineComment}[1]{\State \(\triangleright\) #1}

\title{AEMA: Verifiable Evaluation Framework for Trustworthy and Controlled Agentic LLM Systems}

\author[1]{Yen-Ting Lee}
  \author[2]{Keerthi Koneru}
  \author[3]{Zahra Moslemi}
  \author[2]{Sheethal Kumar}
  \author[2]{Ramesh Radhakrishnan}

  \affil[1]{University of California, San Diego}
  \affil[2]{Center for Advanced AI, Accenture}
\affil[3]{University of California, Irvine}

  \date{\today}
  \correspondingauthor{Keerthi Koneru (\href{mailto:keerthi.koneru@accenture.com}{keerthi.koneru@accenture.com}). All work was done during the authors’ tenure at Accenture.}

\begin{document}
\begin{abstract}
Evaluating large language model (LLM)-based multi-agent systems remains a critical challenge, as these systems must exhibit reliable coordination, transparent decision-making, and verifiable performance across evolving tasks. Existing evaluation approaches often limit themselves to single-response scoring or narrow benchmarks, which lack stability, extensibility, and automation when deployed in enterprise settings at multi-agent scale. We present \textbf{AEMA (Adaptive Evaluation Multi-Agent)}, a process-aware and auditable framework that plans, executes, and aggregates multi-step evaluations across heterogeneous agentic workflows under human oversight. Compared to a single LLM-as-a-Judge, AEMA achieves greater stability, human alignment, and traceable records that support accountable automation. Our results on enterprise-style agent workflows simulated using realistic business scenarios demonstrate that AEMA provides a transparent and reproducible pathway toward responsible evaluation of LLM-based multi-agent systems.

\end{abstract}

\maketitle
\def\Snospace~{Section }
\def\sectionautorefname{\Snospace}
\def\subsectionautorefname{\Snospace}
\def\subsubsectionautorefname{\Snospace}
\def\chapterautorefname{\Snospace}

\section{Introduction}
Agentic LLM systems are moving from static prediction to autonomous reasoning, tool use, and collaboration. As they operate in dynamic environments, their trustworthiness depends on ensuring alignment with human intent, robustness under real-world complexity, and verifiability of their internal decisions. These properties motivate evaluation frameworks that provide auditable transparency and bounded autonomy under human oversight.

Evaluating such systems is difficult because reasoning and coordination unfold over many steps; traditional benchmarks cannot verify process-level reliability or alignment with human expectations.
In high-stakes domains where agents operate under human oversight, evaluation must function as a mechanism of trust and control rather than a mere scoring.

We introduce AEMA (Adaptive Evaluation Multi-Agent), a framework for verifiable evaluation of agentic LLM systems. AEMA adapts to diverse tasks and records traceable evaluation logs for oversight and accountability. Unlike single-model evaluators, AEMA operates as a coordinated multi-agent evaluator that plans, debates, and aggregates judgments across steps to produce consistent and explainable assessments under human control.

To demonstrate feasibility and trust reliability, we evaluate AEMA on enterprise-style agent workflows that simulate realistic multi-agent co-ordination under controlled conditions.

AEMA shows lower score dispersion, stronger human alignment, and verifiable consistency compared to a single LLM-as-a-Judge.

In summary, our contributions advance trust and control in agentic AI through the following dimensions:
\begin{enumerate}
    \item AEMA introduces a process-aware, verifiable framework that unifies step-level and end-to-end assessment for multi-agent evaluation;
    \item Provides an adaptive methodology that offers support for human-in-the-loop oversight and continuous refinement of evaluation criteria;
    \item Demonstrates verifiable evaluation that enables alignment, robustness, and trust in autonomous LLM systems in enterprise-style environments.
\end{enumerate}

\section{Related Work: From Single-Turn Evaluation to Process-Level Assessment of LLM Multi-Agent Systems}

LLMs are increasingly used to automate the evaluation, offering greater scalability and reproducibility than human evaluation. The \textbf{LLM-as-a-Judge} paradigm has been adopted for a wide range of tasks such as instruction~\cite{InstructionFollowing}, summarization~\cite{Summarization}, and machine translation~\cite{MachineTranslation}. Frameworks like G-Eval~\cite{G-Eval} use a single GPT-4 model with a \textbf{chain-of-thought} urging the text to score on coherence, relevance and factual accuracy, while ChatEval~\cite{ChatEval} assembles multiple LLM “referees”, having diverse expertise, which debate to reach consensus. CollabEval~\cite{CollabEval} focuses on building consensus among multiple LLM evaluators using different models rather than adversarial debate. These methods improve consistency, but remain limited to single-turn responses and do not capture multi-step agent reasoning.

Recent research focuses on the evaluation of agentic processes that involve planning, tool use, and interaction. Mind2Web 2~\cite{Mind2Web2} 
benchmarks web agents through an Agent-as-a-Judge rubric that assesses correctness and source attribution in 130 challenging search tasks. WebCanvas~\cite{WebCanvas} measures intermediate actions in realistic interface states using an online evaluation framework, and BEARCUBS~\cite{BEARCUBS} provides a multimodal benchmark of browsing tasks to test the effectiveness of reasoning.

Another line of work provides broad benchmark suites to measure how well modern LLM-driven agents perform in specialized domains. MLAgentBench~\cite{MLAgentBench} introduces a benchmark of 13 end-to-end machine learning experimentation tasks that evaluate agents ability to plan, write code, run experiments, and analyze results. Similarly, ITBench~\cite{ITBench} targets real-world IT automation tasks in Site Reliability Engineering (SRE), Compliance/Security (CISO), and Financial Operations (FinOps).

Beyond outcome accuracy, several studies assess behavior quality and collaboration dynamics.
AutoLibra~\cite{AutoLibra} derives metrics from clustered human feedback for  fine-grained behavior assessment. GEMMAS~\cite{GEMMAS} introduces graph-based metrics that measure 
collaboration efficiency and redundancy in multi-agent reasoning. WorFEval~\cite{WorFEval} compares 
workflow graphs against ground-truth structures using subsequence and subgraph matching, while 
DevAI~\cite{DevAI} monitors the progress of another agent to provide step level feedback on coding tasks.

These studies collectively advance LLM-based evaluation, but primarily assess single-response quality, benchmarked outcomes, or isolated collaboration patterns. In contrast, our AEMA evaluation framework unifies process-level planning, adaptive scoring, and historical trace learning into a reproducible evaluation workflow suited for enterprise-style multi-agent systems.
\section{Method}
\subsection{AEMA Framework}

\begin{figure}[t]
  \centering
  \includegraphics[width=0.9\linewidth, keepaspectratio]{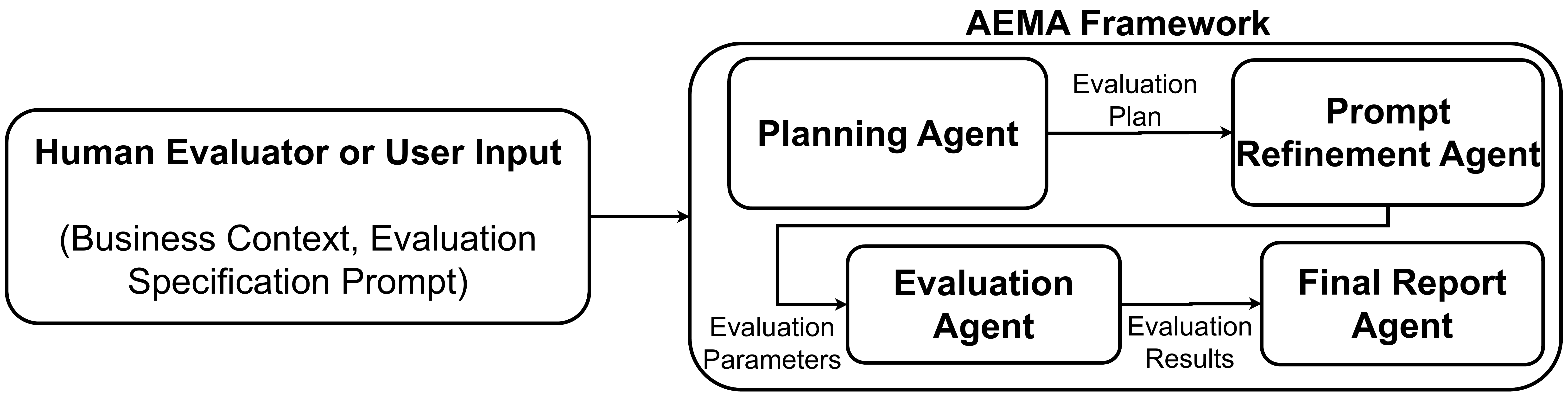}
  \caption{\textbf{Overview of AEMA (Adaptive Evaluation Multi-Agent):} Planning Agent builds the plan; Prompt-Refinement Agent retrieves and prepares examples; Evaluation Agents score intermediate actions; Final Report Agent aggregates results into an auditable, reproducible report.}
  \label{fig:Framework}
\end{figure}

\textbf{AEMA} (Adaptive Evaluation Multi-Agent) is a verifiable multi-agent evaluation loop with four roles: Planning, Prompt-Refinement, Evaluation, and Final Report.
It analyzes execution traces and criteria to plan, parameterize, and score step-wise behaviors, then aggregates auditable reports under human oversight Figure~\ref{fig:Framework}. This \textbf{multi role workflow creates a controlled evaluation loop} that records every decision for auditability and post-hoc verification.

\subsection{Planning Agent}\label{sec:planning}

The Planning Agent determines which tasks to evaluate based on the evaluation prompt specified by the human evaluator. By default, it evaluates all actions, but prioritizes those that are semantically meaningful within the workflow. In complex multi-agent systems, agents often engage in extended discussions or reasoning to clarify objectives. As shown in GEMMAS~\cite{GEMMAS}, naive agent pipelines tend to generate high redundancy and low diversity in such internal communication. To avoid unnecessary overhead, the Planning Agent excludes these intermediary steps from the evaluation and instead focuses on key responses, tool calls, and actions assessable by available functions.

To support AEMA in various business scenarios and domains, the Planning Agent first classifies the business context into a predefined category using an LLM that analyzes the agent’s execution results and the evaluation prompt. 
By assigning the task to a user-specified business category, the Planning Agent ensures that subsequent evaluations are aligned with the domain. This process also reduces the complexity of selecting the appropriate evaluation functions by restricting the search space to those relevant within the identified business domain.

Even within the same business domain, the large number of possible actions makes it difficult for the Planning Agent to reliably select the correct evaluation functions. To address this challenge, we introduce a tool filter that improves the efficiency of retrieving relevant evaluation functions. 
 
The filter retrieves only functions aligned with the identified domain and evaluator requirements. Inspired by Retrieval-Augmented Generation (RAG)~\cite{RAG}, we adopt a hybrid retrieval approach that combines sparse and dense search methods over the function descriptions (docstrings). Combining sparse and dense retrieval methods has been empirically shown to yield more accurate and robust results than using either alone, particularly in domain-specific applications such as scientific document retrieval~\cite{Retrieval}. In our setting, we used the sparse retrieval implemented by BM25 to capture keyword level relevance. For dense retrieval, we compute the cosine similarity between the embeddings of the function descriptions and the embeddings of the task context, which includes both the agent’s execution results and the evaluation prompt. Finally, we use a convex combination of sparse and dense scores to rank the functions and select the top-k candidates (where k is set by the human evaluator). This filtering step enables the Plan Generator to focus on the most appropriate evaluation tools for the given business scenario.

After retrieving relevant functions, the Planning Agent combines two components: a Plan Generator and a Plan Evaluator. Both use the same underlying model (GPT-4o) but with different system messages and prompts. The Plan Generator selects which steps of the business agent should be evaluated. After producing a candidate plan, the Plan Evaluator reviews it. If both reach consensus, the plan proceeds; otherwise, the Plan Evaluator provides feedback, prompting the Plan Generator to revise. This iterative process continues until agreement or a maximum iteration limit is reached.

\subsection{Prompt Refinement Agent}

For domain-specific tasks, pretrained LLMs may be insufficient to reliably evaluate a business agent without contextual grounding. Previous work \cite{ManyShotJudge} shows that LLM-as-a-Judge achieves higher accuracy when evaluation prompts include few-shot exemplars rather than zero-shot queries. To support this, AEMA’s Prompt-Refinement Agent automatically searches prior evaluation records relevant to the current task; when suitable examples are missing, it synthesizes exemplars aligned to the task of the surrounding business context.

The Prompt Refinement Agent prepares the required parameter bundle and few-shot example for each evaluation function in the plan generated by the Planning Agent. 
It uses an LLM to interpret the signature and docstring of each function along with the agent’s inputs, outputs, and reasoning trace, mapping relevant facts to arguments. The Prompt Refinement Agent produces a schema-compliant JSON object for immediate use by the evaluation function.

In-context learning research, such as Auto-ICL~\cite{Auto-ICL} and Self-ICL~\cite{Self-ICL}, shows that LLMs perform better when prompts include clear instructions and examples. To support such evaluation functions, the Prompt-Refinement Agent automatically retrieves past evaluation results using the same hybrid retrieval method described in~\ref{sec:planning} 
When an evaluation function requires few-shot examples but the database does not provide enough, the agent asks an LLM to synthesize examples from the business context, including example agent actions and reference scores. 
This automated extraction and adaptive retrieval improve flexibility when extending evaluation functions.

We implement AEMA using AutoGen \cite{AutoGen}, which provides asynchronous orchestration and memory integration through ChromaDB and LangChain capabilities beyond lighter frameworks such as the OpenAI Agents SDK and CrewAI. Our deployment uses GPT-4o as the underlying LLM and ChromaDB as the vector store.

\subsection{Evaluation Agents}

The Evaluation Agents consist of multiple general evaluators capable of assessing each task performed by the business agent. They measure individual skills using both LLM-based judges and reliable code-based evaluators. Each evaluation function produces a normalized score between zero and one based on the evaluation results, along with qualitative feedback for further analysis.

Each function outputs a normalized score (0–1) and qualitative feedback. After all evaluations, results are sent to the \textbf{Final Report Agent} to generate the final report summarizing key metrics and improvement areas. Although the framework supports domain-specific evaluation extensions, we also provide a set of general functions that are applicable to all business scenarios.

\subsection{Final Report Agent}

The Final Report Agent consolidates all intermediate results into a comprehensive performance report for the target business agent. It aggregates quantitative metrics and qualitative insights produced by the Evaluation Agents into a coherent summary.
 
The resulting report typically consists of three elements: 
\begin{itemize}
    \item  an aggregated performance score, obtained by averaging or weighting across all available evaluation dimensions;
    \item a performance summary organizing evaluator feedback into strengths and weaknesses, and 
    \item actionable recommendations, which highlight concrete areas for improvement and suggest strategies to enhance the business agent’s reliability.
\end{itemize}

The Final Report Agent also supports continuous monitoring beyond static assessment. By recording and analyzing historical evaluation trajectories, it detects performance trends and distribution shifts in evolving business data. This functionality ensures robustness in dynamic environments. This design enables accountable and auditable evaluation traces, a prerequisite for responsible multi-agent deployment.

\section{Experiments}
\label{sec:experiments}

This section evaluates the effectiveness of AEMA in generating reliable, human-aligned assessments in enterprise-style multi-agent workflows. We describe the experimental setup, domain configuration, task design, and then analyze the results to highlight stability, alignment, and responsible evaluation outcomes.

\subsection{Experiment Settings}


To demonstrate the operation of AEMA, we evaluate a finance domain agent responsible for verifying the validity of the invoice within a multi-agent workflow. The workflow involves six coordinated components: InputAgent (analyzes the incoming and normalizes input), Orchestrator (plans the collaboration order among agents), ParserAgent (extracts information from the invoice), Validator Agent (validates the extracted results), Policy Agent (checks compliance with policies), and Approval Agent (aggregates the outcomes and gives approval decision), each corresponding to a functional stage in enterprise automation. Although this study focuses on finance, AEMA supports domain-specific evaluation functions across Finance, Healthcare, and Software, automatically restricting its plan to the detected domain during planning.

AEMA dynamically configures its evaluation plan according to available functions. To test its adaptive planning capability, we intentionally excluded the evaluation of the Policy Agent while keeping others active, requiring AEMA to generate a coherent plan reflecting only implementable checks.

To assess the reliability of plan generation, AEMA was executed 30 times on identical financing workflow output. Within the Planning Agent, we capped the debate mechanism at five rounds for both the Plan Generator and the Plan Evaluator. This configuration isolates the contribution of debate rounds to planning accuracy and convergence speed.


For consistency testing, both AEMA and the single LLM-as-a-Judge baseline were executed 30 times on identical inputs. To ensure fairness, the single LLM used concatenated scoring criteria identical to AEMA evaluation functions. The results were benchmarked against human-as-a-Judge references to measure alignment and dispersion.

We evaluated two input conditions: clear (good quality) and degraded (blurry) invoice images. For each, AEMA and the single LLM-as-a-Judge were run 20 times, and their mean scores were compared to human references to evaluate robustness under input quality shifts.

\subsection{Formal Plan-Evaluation Metric}

Let the available agent set be~$\Omega$.
The gold plan is $G_i=(A_i,S_i)$ with $A_i\subseteq\Omega$ and order $S_i$.
The prediction is $(P,E)$ with $P\subseteq\Omega$ and predicted sequence $E$.
AEMA returns a scalar score in $[0,1]$ summarizing five criteria:

\begin{enumerate}
  \item \textbf{Schema validity ($F$).}
  Fraction of required fields that are present with the correct types, ensuring structurally valid outputs.

  \item \textbf{Agent selection accuracy ($S$).}
  The $F_1$-score between the gold and predicted agent sets:
  \[
  S=\frac{2\,|A_i\cap P|}{|A_i|+|P|}.
  \]

  \item \textbf{Step--agent coherence ($C_v$).}
  Measures the overlap between the unique step labels and the predicted agents:
  \[
  C_v=\frac{|\mathrm{set}(E)\cap P|}{|\mathrm{set}(E)\cup P|}.
  \]

  \item \textbf{Order preservation ($O_r$).}
  Assesses whether the predicted sequence $E$ preserves the intended order~$S_i$.  
  Using a Kendall-style inversion agreement~\cite{cicirello2019kendall}:
  \[
  O_r = 1-\frac{\mathrm{inv}(E;S_i)}{\mathrm{inv}_{\max}}, \qquad O_r\in[0,1],
  \]
  where $\mathrm{inv}(E;S_i)$ is the number of pairwise inversions of $E$ relative to~$S_i$;
  for DAGs, inversions are computed with respect to any consistent topological extension.

  \item \textbf{Step efficiency ($E_f$).}
  Encourages parsimony in plan length:
  \[
  E_f = \min\!\left(1,\;\frac{|A_i|}{\max(1,|P|)}\right).
  \]
\end{enumerate}

\noindent\textbf{Aggregation via AHP.}  
A non-negative weight vector $(w_F,w_S,w_{C_v},w_{O_r},w_{E_f})$ with $\sum w_\cdot=1$
is obtained using the \emph{Analytic Hierarchy Process (AHP)}~\cite{bhattacharya2017granger,botchway2021ahpsoftware,mekouar2021ahpspam}
from pairwise criterion comparisons with a consistency check.
The final score is:
\[
\mathrm{\textit{FinalScore}} = 
w_F F + w_S S + w_{C_v} C_v + w_{O_r} O_r + w_{E_f} E_f.
\]
Weights can be adjusted to reflect domain priorities without altering metric definitions.

\medskip
\noindent\textbf{Notation Summary.}  
$\Omega$: all candidate agents; 
$A_i$: gold-standard agent set; 
$S_i$: gold order / DAG; 
$P$: predicted agent set; 
$E$: predicted sequence; 
$\mathrm{inv}(\cdot)$: inversion count used in order-preservation.

\subsection{Results and Analysis}
\paragraph{Planning Experiment.}

We evaluated AEMA’s Planning Agent over 30 runs. The consensus on the correct plan was reached in 13 runs after one debate round, 13 runs after two rounds, and 4 runs after three rounds. This supports using a debate mechanism with Plan Generator and Plan Evaluator in Planning Agent, since a single LLM generation is not consistently optimal. In this setup, a cap of three rounds is sufficient. 

\paragraph{Stability Experiment.}

\begin{figure}[t]
  \centering
  \includegraphics[width=0.85\linewidth, keepaspectratio]{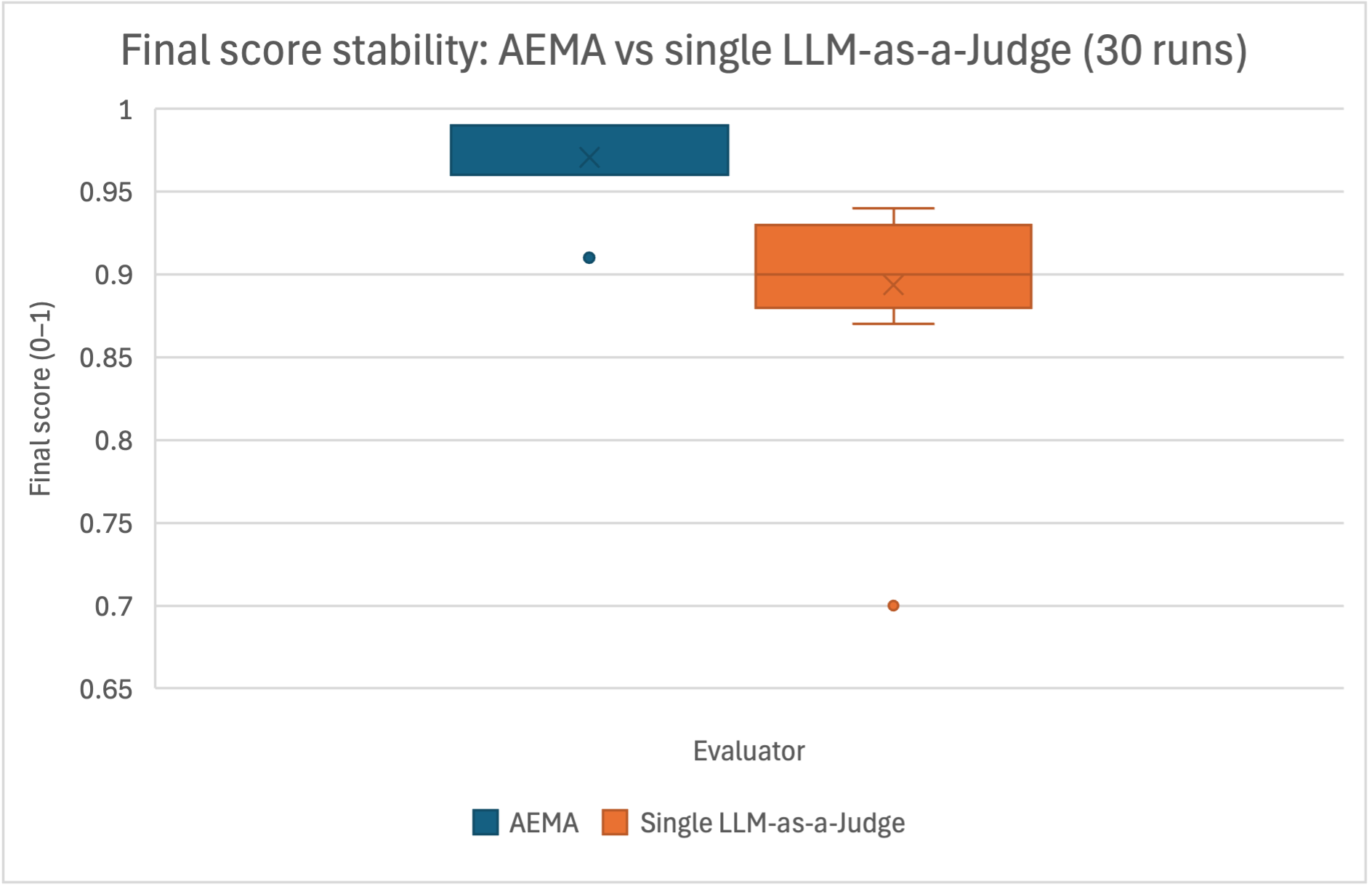}
  \caption{Stability of final scores across 30 evaluations. AEMA exhibits lower dispersion than a single LLM-as-a-Judge.}
  \label{fig:Stability}
\end{figure}


Across 30 evaluations with identical inputs, AEMA shows narrow score variation at each step and in the Final metric. In contrast, the single LLM-as-a-Judge shows wider dispersion with several low outliers, especially for Decision and Final. The box-and-whisker plots in Figure~\ref{fig:Stability} show a clearly smaller interquartile range and shorter whiskers for AEMA. The human-as-a-judge reference for this case is 0.96, and AEMA’s distribution concentrates near this value, while the single-LLM distribution is broader and departs more frequently from it. In business workflows, stability is essential. Predictable evaluations support fixed thresholds, reduce human overrides, and improve auditability and policy compliance. Under fixed inputs, these results show that AEMA yields more stable and human-aligned assessments than a single LLM-as-a-Judge. In high-reliability settings, this consistency reduces the evaluator drift and enables formal threshold calibration.

\paragraph{Human Alignments Experiment.}

\begin{table}[t]
\centering
{\small
\setlength{\tabcolsep}{4pt}
\renewcommand{\arraystretch}{1.1}
\begin{tabular}{l c cc cc}
\hline
 & Human & \multicolumn{2}{c}{AEMA} & \multicolumn{2}{c}{Single LLM-as-a-Judge} \\
\cline{3-4}\cline{5-6}
Step & Mean & Mean & $|\Delta|$ & Mean & $|\Delta|$ \\
\hline
Routing    & 1.00 & 0.98 & 0.02 & 1.00 & 0.00 \\
Planning   & 0.98 & 0.98 & 0.00 & 0.86 & 0.12 \\
Parsing    & 0.80 & 0.88 & 0.08 & 0.86 & 0.06 \\
Validation & 1.00 & 1.00 & 0.00 & 1.00 & 0.00 \\
Decision   & 1.00 & 1.00 & 0.00 & 0.81 & 0.19 \\
Final      & 0.96 & 0.97 & 0.01 & 0.87 & 0.09 \\
\hline
\end{tabular}
}
\caption{Good-quality invoice: mean scores (over 20 runs) and absolute error to human (lower is better).}
\label{tab:good-align}
\end{table}

Table 1 shows the good-quality invoice results. AEMA matches the human score on planning, validation and decision and is close to the final score (absolute error = 0.01). The single LLM-as-a-Judge is farther from the human score, especially in the decision step (0.19) and the final (0.09). Averaging the absolute error over the six steps, the AEMA is 0.018, while the single LLM is 0.077.

Table 2 shows the results of the degraded and blurred input images. AEMA remains closer to the human score on most steps and on the final score (0.04 vs. 0.07). The single LLM departs more on Decision (0.32) and Validation (0.12). The average absolute error over steps is 0.037 for AEMA and 0.108 for the single LLM.

Overall, AEMA is more human-aligned under both input conditions. The greatest gains occur in the decision step and the final score, which determine the business outcome.

\begin{table}[t]
\centering
{\small
\setlength{\tabcolsep}{4pt}
\renewcommand{\arraystretch}{1.1}
\begin{tabular}{l c cc cc}
\hline
 & Human & \multicolumn{2}{c}{AEMA} & \multicolumn{2}{c}{Single LLM-as-a-Judge} \\
\cline{3-4}\cline{5-6}
Step & Mean & Mean & $|\Delta|$ & Mean & $|\Delta|$ \\
\hline
Routing    & 1.00 & 0.94 & 0.06 & 1.00 & 0.00 \\
Planning   & 1.00 & 1.00 & 0.00 & 0.99 & 0.01 \\
Parsing    & 0.50 & 0.38 & 0.12 & 0.63 & 0.13 \\
Validation & 1.00 & 1.00 & 0.00 & 0.88 & 0.12 \\
Decision   & 1.00 & 1.00 & 0.00 & 0.68 & 0.32 \\
Final      & 0.90 & 0.86 & 0.04 & 0.83 & 0.07 \\
\hline
\end{tabular}
}
\caption{Blurry invoice: mean scores (over 20 runs) and absolute error to human (lower is better).}
\label{tab:blurry-align}
\end{table}


\section{Limitations and Future Directions}

\textbf{Score Stability of Continuous Outputs.}  
Repeated evaluations of identical parsing steps occasionally produce inconsistent numerical scores despite coherent rationales,
highlighting variance in continuous-value judgments.
Discrete or categorical scoring scales are recommended to improve repeatability and interpretability.

\medskip
\noindent\textbf{Evaluation Cost and Latency.}  
AEMA involves multiple LLM calls (\textit{plan} $\rightarrow$ \textit{prompt} $\rightarrow$ \textit{evaluation} $\rightarrow$ \textit{report}),
which can be costly and slower than the business workflow itself.
Practical mitigations include evaluating only priority actions, skipping redundant re-evaluations within a time window, and using budget-aware planning to allocate small versus large models dynamically.
Caching repeated prefixes (e.g., KV-cache reuse) further reduces runtime overhead.

\medskip
\noindent\textbf{Prompt-Merging Trade-offs.}  
Combining stages (such as parameter extraction and evaluation) reduces the number of calls,
but early observations suggest that heavily merged prompts may lower task accuracy or interpretability.
Controlled experiments are planned to identify safe levels of consolidation without compromising transparency.

\medskip
\noindent\textbf{Scope and Generalization.}  
Current experiments are confined to a single Finance domain using LLM-based components across all agents.
Future extensions will explore cross-domain settings, including safety-critical workflows, and investigate cost and latency reductions via
discrete scoring, caching, and optimized prompt compilation.

\section{Conclusion}
We presented AEMA, a process-aware evaluation framework for business agents. AEMA demonstrates that evaluation can serve as a verifiable control mechanism embedding accountability into multi-agent reasoning. AEMA plans evaluations, prepares parameters and examples, applies multiple evaluators, and aggregates step-level and end-to-end results. Compared with a single LLM-as-a-Judge, AEMA produces more stable scores under identical inputs and yields smaller absolute error to human references on both good-quality and degraded invoices. In business settings, AEMA yields stable and predictable scores that support fixed thresholds and reduce manual overrides. 

Although this study focuses primarily on the Finance domain, AEMA’s architecture demonstrates strong cross-domain adaptability. Future work will extend domain-specific evaluation experiments to broader settings, testing AEMA across multiple domains and heterogeneous workflows. Another promising direction is to improve the efficiency and awareness of AEMA resources. For example, integrating cost or budget constraints into the workflow could allow the Planning Agent to activate evaluation tools dynamically based on available resources or to decide when to use small language models versus larger ones to balance accuracy and efficiency. These extensions would make AEMA more scalable and practical for real-world, cost-sensitive applications.
\newpage

\raggedright
\bibliographystyle{iclr2025_conference}
\bibliography{contents/references}

\newpage
\appendix
\section{System \& Workflow Summary}

AEMA evaluates multi-agent business workflows through four coordinated roles:
\begin{enumerate}
    \item \textbf{Planning Agent} --- generates an executable evaluation plan from the human evaluator’s specification and relevant evaluation functions;
    \item \textbf{Prompt-Refinement Agent} --- prepares evaluator parameters and few-shot exemplars for each function;
    \item \textbf{Evaluation Agents} --- combine deterministic checks and LLM-based judgments to score each step; and
    \item \textbf{Final Evaluation Agent} --- aggregates outcomes into machine- and human-readable reports.
\end{enumerate}

Each stage contributes to a reproducible, auditable evaluation trace.
Detailed implementation of these roles, the plan–review loop, and retrieval mechanisms appear in Section~3 of the main text.
Here we only note that the framework is implemented in \textbf{AutoGen}, integrated with \textbf{LangChain} and \textbf{ChromaDB}, and uses \textbf{GPT-4o} as the LLM backbone.

\vspace{3mm}
\noindent\textit{System design principle:}
AEMA ensures that every evaluation decision includes plan creation, parameter refinement, scoring, and reporting, which produces a verifiable artifact that can be replayed or audited later,
reinforcing the workshop theme of \textbf{trust and control in agentic AI}.

\section{Experimental Configuration}

\textbf{Target Workflow.}  
The \textbf{Finance Agent Framework} validates invoices through six components:
\textit{InputAgent}, \textit{Orchestrator}, \textit{ParserAgent}, \textit{ValidatorAgent}, \textit{PolicyAgent}, and \textit{ApprovalAgent}.
This workflow was selected to emulate realistic enterprise evaluation flows involving document parsing, rule validation, and policy compliance.

\medskip
\noindent\textbf{Evaluator-Availability Stress Test.}  
To test adaptive planning, the evaluation for \textit{PolicyAgent} is intentionally omitted while all other component evaluations remain implemented.
AEMA must construct a coherent evaluation plan reflecting the available evaluators without failing the planning step.

\medskip
\noindent\textbf{Planning-Stability Protocol.}  
AEMA is executed 30~times on identical Finance-agent outputs.
The Planning Agent’s debate loop is capped at five~rounds for both the Plan Generator and the Plan Evaluator.
Consensus on the correct plan typically emerges after one to three~rounds (13, 13, and 4~runs respectively), showing bounded deliberation and stable reasoning behavior.

\medskip
\noindent\textbf{Stability Under Identical Inputs.}  
Across 30~repeated evaluations on the same inputs, AEMA yields narrow score dispersion at both step and final levels,
while a single LLM-as-a-Judge baseline exhibits larger variance and occasional outliers.
Human-as-a-judge references cluster near~0.96, and AEMA’s scores concentrate around this value.
These results demonstrate robustness of AEMA’s debate-and-review mechanism under fixed conditions (refer Figure~\ref{fig:Stability}).

\medskip
\noindent\textbf{Human-Alignment Evaluation.}  
Two datasets are used: good-quality invoices and degraded (blurred) images.
Both AEMA and the single-LLM baseline are run twenty~times per dataset, and the averaged results are compared to human-judge references.
For good-quality inputs, AEMA aligns closely with human ratings across all steps, while the single LLM diverges more strongly in the \textit{Decision} and \textit{Final} stages.
Under degraded inputs, AEMA maintains smaller absolute deviations (see Tables~\ref{tab:good-align}--\ref{tab:blurry-align}), indicating resilience to input noise.

\medskip
\noindent\textbf{Baseline Fairness.}  
For the single LLM-as-a-Judge baseline, the same scoring criteria prompts used in AEMA’s evaluation functions are concatenated into one unified prompt to ensure a fair comparison under identical input conditions.

\end{document}